\documentclass[sigconf]{acmart}
\AtBeginDocument{%
  }

\usepackage{amsmath} 
\usepackage{subcaption}
\usepackage{algorithm}
\usepackage{algorithmic}
\usepackage{multirow} 
\usepackage{tabularx} 
\usepackage{pifont}
\usepackage{booktabs}

\newcommand{\cmark}{\ding{51}}
\newcommand{\xmark}{\ding{55}}

\copyrightyear{2025}
\acmYear{2025}
\setcopyright{acmlicensed}\acmConference[WWW '25]{Proceedings of the ACM Web Conference 2025}{April 28-May 2, 2025}{Sydney, NSW, Australia}
\acmBooktitle{Proceedings of the ACM Web Conference 2025 (WWW '25), April 28-May 2, 2025, Sydney, NSW, Australia}
\acmDOI{10.1145/3696410.3714836}
\acmISBN{979-8-4007-1274-6/25/04}

\begin{document}

\title{CTR-Driven Advertising Image Generation with Multimodal Large Language Models}

\settopmatter{authorsperrow=4}

\author{Xingye Chen}
\authornote{Work done while interning at JD.COM.}
\authornote{Both authors contributed equally to this research.}
\affiliation{%
  \institution{Huazhong University of Science and Technology}
  \city{Wuhan}
  \country{China}}
\email{chenxingye@hust.edu.cn}
\orcid{0000-0002-7119-4792}

\author{Wei Feng}
\authornotemark[2]

\affiliation{%
  \institution{JD.COM}
  \city{Beijing}
  \country{China}
}
\email{fengwei25@jd.com}
\orcid{0009-0005-8890-4956}

\author{Zhenbang Du}
\authornotemark[1]

\affiliation{%
  \institution{Huazhong University of Science and Technology}
  \city{Wuhan}
  \country{China}
}
\email{dzb99@hust.edu.cn}
\orcid{0000-0002-1386-8381}

\author{Weizhen Wang}
\affiliation{%
  \institution{JD.COM}
  \city{Beijing}
  \country{China}
}
\email{wangweizhen5@jd.com}
\orcid{0009-0001-4006-774X}

\author{Yanyin Chen}
\affiliation{%
  \institution{JD.COM}
  \city{Beijing}
  \country{China}
}
\email{chenyanyin6@jd.com}
\orcid{0009-0004-5508-3605}

\author{Haohan Wang}
\affiliation{%
  \institution{JD.COM}
  \city{Beijing}
  \country{China}
}
\email{wanghaohan1@jd.com}
\orcid{0000-0003-3451-6884}

\author{Linkai Liu}
\authornotemark[1]
\affiliation{%
  \institution{Sun Yat-sen University}
  \city{Shenzhen}
  \country{China}
}
\email{liulk6@mail2.sysu.edu.cn}
\orcid{0000-0003-3748-5980}

\author{Yaoyu Li}
\affiliation{%
  \institution{JD.COM}
  \city{Beijing}
  \country{China}
}
\email{liyaoyu1@jd.com}
\orcid{0000-0002-7362-7897}

\author{Jinyuan Zhao}
\affiliation{%
  \institution{JD.COM}
  \city{Beijing}
  \country{China}
}
\email{zhaojinyuan1@jd.com}
\orcid{0000-0001-8528-3529}

\author{Yu Li}
\affiliation{%
  \institution{JD.COM}
  \city{Beijing}
  \country{China}
}
\email{liyu1078@jd.com}
\orcid{0000-0003-1331-5020}

\author{Zheng Zhang}
\affiliation{%
  \institution{JD.COM}
  \city{Beijing}
  \country{China}
}
\email{zhangzheng11@jd.com}
\orcid{0009-0002-6391-4814}

\author{Jingjing Lv}
\affiliation{%
  \institution{JD.COM}
  \city{Beijing}
  \country{China}
}
\email{lvjingjing1@jd.com}
\orcid{0009-0000-5518-7077}

\author{Junjie Shen}
\affiliation{%
  \institution{JD.COM}
  \city{Beijing}
  \country{China}
}
\email{shenjunjie@jd.com}
\orcid{0009-0008-6983-5213}

\author{Zhangang Lin}
\affiliation{%
  \institution{JD.COM}
  \city{Beijing}
  \country{China}
}
\email{linzhangang@jd.com}
\orcid{0000-0003-1379-5044}

\author{Jingping Shao}
\affiliation{%
  \institution{JD.COM}
  \city{Beijing}
  \country{China}
}
\email{shaojingping@jd.com}
\orcid{0000-0001-8555-2020}

\author{Yuanjie Shao}
\authornote{Corresponding author.}
\affiliation{%
  \institution{Huazhong University of Science and Technology}
  \city{Wuhan}
  \country{China}
}
\email{shaoyuanjie@hust.edu.cn}
\orcid{0000-0003-1141-0454}

\author{Xinge You}
\affiliation{%
  \institution{Huazhong University of Science and Technology}
  \city{Wuhan}
  \country{China}
}
\email{youxg@hust.edu.cn}
\orcid{0000-0002-6227-1346}

\author{Changxin Gao}
\affiliation{%
  \institution{Huazhong University of Science and Technology}
  \city{Wuhan}
  \country{China}
}
\email{cgao@hust.edu.cn}
\orcid{0000-0003-2736-3920}

\author{Nong Sang}
\affiliation{%
  \institution{Huazhong University of Science and Technology}
  \city{Wuhan}
  \country{China}
}
\email{nsang@hust.edu.cn}
\orcid{0000-0002-9167-1496}

\renewcommand{\shortauthors}{Xingye Chen, et al.}

\begin{abstract}
In web data, advertising images are crucial for capturing user attention and improving advertising effectiveness. 
Most existing methods generate background for products primarily focus on the aesthetic quality, which may fail to achieve satisfactory online performance.
To address this limitation, we explore the use of Multimodal Large Language Models (MLLMs) for generating advertising images by optimizing for Click-Through Rate (CTR) as the primary objective.
Firstly, we build targeted pre-training tasks, and leverage a large-scale e-commerce multimodal dataset to equip MLLMs with initial capabilities for advertising image generation tasks.
To further improve the CTR of generated images, we propose a novel reward model to fine-tune pre-trained MLLMs through Reinforcement Learning (RL), which can jointly utilize multimodal features and accurately reflect user click preferences.
Meanwhile, a product-centric preference optimization strategy is developed to ensure that the generated background content aligns with the product characteristics after fine-tuning, enhancing the overall relevance and effectiveness of the advertising images.
Extensive experiments have demonstrated that our method achieves state-of-the-art performance in both online and offline metrics. Our code and pre-trained models are publicly available at: \href{https://github.com/Chenguoz/CAIG}{https://github.com/Chenguoz/CAIG}.
\end{abstract}




\maketitle

\section{Introduction}

\begin{figure*}[t]
    \centering
    \includegraphics[width=0.95\textwidth]{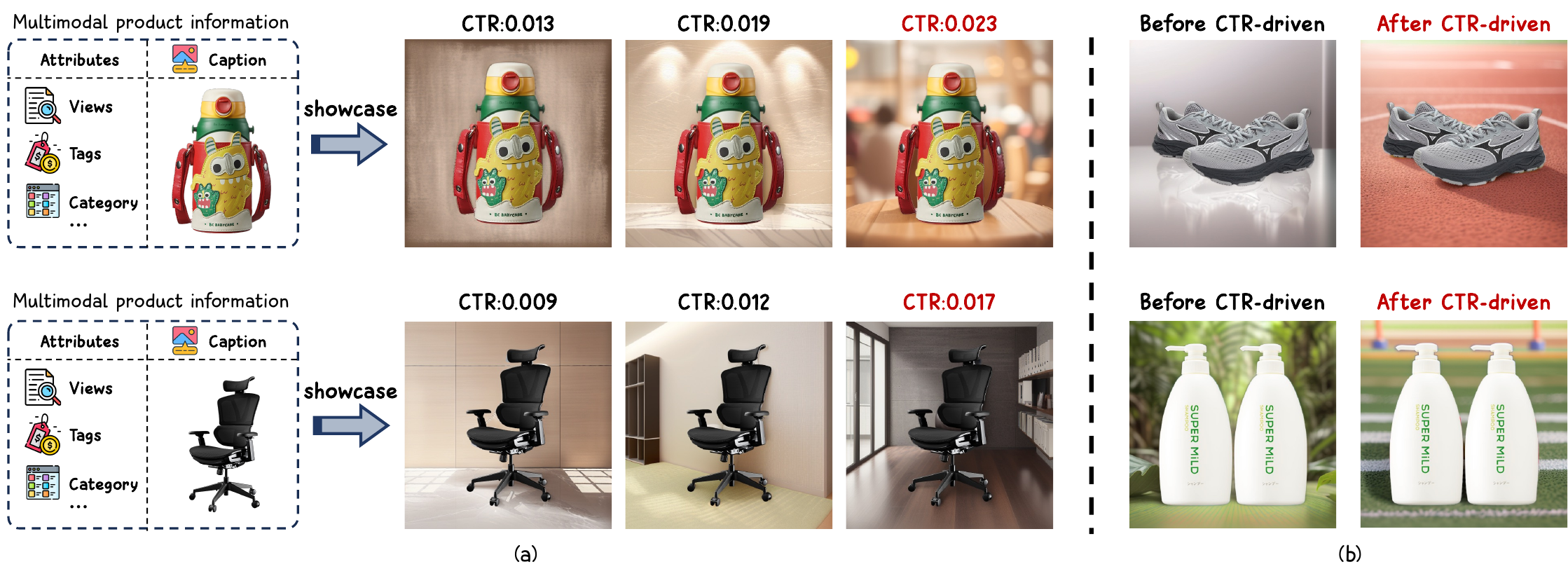}
    \vspace{-1em}
    \caption{(a) Example of the impact of different backgrounds on product CTR. While visual features play a crucial role, other modalities such as textual caption and product attributes also have a significant influence on CTR. (b) Examples of product-background mismatches using existing reinforcement learning algorithms.}
    \label{fig:introduction}
\end{figure*}

Advertising images play a pivotal role in attracting user attention and  boosting advertising efficacy~\cite{mishra2020learning,ku2023staging}.
Recent advancements in image generation techniques, particularly the integration of Stable Diffusion~\cite{rombach2022high} and ControlNet~\cite{zhang2023adding}, have enabled the creation of harmonious and realistic backgrounds for product images. 
However, most existing advertising image generation approaches~\cite{wang2022creagan,zhao2024enhancing,wang2025generate,li2023relation} primarily focus on offline metrics, such as image quality or semantic consistency, without fully considering the critical connection between visual content and online performance metrics like Click-Through Rate (CTR). This results in a notable discrepancy between the generated advertising images and the ideal images that align with actual user preferences.

Inspired by recent approaches~\cite{wei2022creater,yang2024new,lee2024parrot, du2025towards} that incorporate Reinforcement Learning from Human Feedback (RLHF)~\cite{macglashan2017interactive,christiano2017deep,stiennon2020learning} to align with human preferences, we can adopt a two-stage method to better capture online user preferences.
The first stage involves collecting and analyzing online user feedback to train a Reward Model (RM) that accurately simulates user preferences in the e-commerce domain. In the second stage, we employ Reinforcement Learning (RL) algorithms to fine-tune the generation model, with the RM providing rewards to guide the optimization process. A critical aspect of this pipeline is the RM's ability to accurately reflect users' click preferences for images. However, previous methods that incorporate visual content for CTR prediction face two major limitations:
First, applying different backgrounds to the same product can lead to significantly different CTR outcomes, as illustrated in Figure~\ref{fig:introduction} (a). Existing methods~\cite{ge2018image,wang2021hybrid, yang2024new,lin2022joint} often rely on models with limited image understanding capabilities, such as CNNs, vision transformers, or embedding-based methods. To compensate for this deficiency, these methods typically require incorporating numerous auxiliary tasks, such as object detection and OCR, which leads to additional annotation costs and labor-intensive data preparation processes.
Second, integrating diverse yet crucial features from multiple modalities (such as product titles and attributes) is of paramount importance, as these significantly influence product CTR. Nevertheless, current methods primarily focus on dense visual features and require additional complex modules to fuse different types of features, potentially limiting the model's adaptability to the rapidly changing online advertising environments. For instance, products from distinct categories, such as water bottles and office chairs illustrated in Figure~\ref{fig:introduction} (a), exhibit remarkably different baseline CTR due to their disparate nature and associated consumer behavior patterns.

To address these issues, leveraging the advanced multimodal understanding and representation capabilities of MLLMs \cite{liu2024visual, liu2024llavanext, yang2024qwen2} offers a promising solution.
On the one hand, these models excel in zero-shot visual analysis, encompassing image representation~\cite{liu2023mllms, jain2024vcoder}, object detection~\cite{li2024llava, zang2024contextual}, and various visual tasks without requiring task-specific training.  On the other hand, by transforming sparse features (such as categories, tags, or other attributes)  into natural language descriptions, MLLMs can process and reason about this textual information alongside visual data, offering a simpler paradigm for integrating multimodal information.
While the introduction of MLLMs can effectively guide generation models to produce backgrounds with higher CTR, it is crucial to consider the relationship between the background and the product in advertising image generation. Existing RL algorithms~\cite{wu2023better,lee2023aligning,lee2024parrot} focus solely on optimizing rewards, neglecting the crucial balance between visual appeal and contextual appropriateness. This oversight can result in disharmonious backgrounds that mislead users and lead to poor shopping experiences.
As illustrated in Figure \ref{fig:introduction} (b), while dynamic, sports-oriented backgrounds might boost CTR for athletic shoes, the model might erroneously apply similar backgrounds to unrelated products like cosmetics, compromising visual harmony and product relevance.

In this work, we propose a novel method called \textbf{C}TR-driven \textbf{A}dvertising \textbf{I}mage \textbf{G}eneration (CAIG), which leverages the MLLMs  as core components to generate advertising images that are both CTR-optimized and coherent with product characteristics.
As illustrated in Figure~\ref{fig:1}, we first design targeted pre-training tasks that utilize a large-scale e-commerce multimodal dataset to equip MLLMs with comprehensive e-commerce domain knowledge for generating advertising images. 
To further optimize the CTR of generated images, we propose a novel RM that transforms the traditional CTR prediction task into a binary classification problem, enabling the selection of positive and negative samples in subsequent RL process. By focusing on the relative performance between image pairs, our method can effectively mitigate the impact of absolute CTR variations across different product categories.
Lastly, to avoid generating background-irrelevant advertisement images, we develop a Product-Centric Preference Optimization (PCPO) strategy. This strategy uses multimodal information of the product as the sole variable and constructs additional preference pairs, forcing the MLLM to generate background content that aligns with the product's characteristics during the RL process.
To the best of our knowledge, this is the first work that utilizes MLLMs for CTR-driven advertising image generation.

We summarize our contributions as three-folds:
\vspace{-0.3em}
\begin{itemize}
	\item
We design targeted pre-training tasks using a large-scale e-commerce multimodal dataset to equip MLLMs with comprehensive domain knowledge, providing them with foundational capabilities for downstream tasks.
	\item
We propose a two-branch RM that combines the powerful image understanding capabilities of MLLMs with multimodal product information fusion to effectively simulate human click preferences in e-commerce scenarios.
	\item
We develop a product-centric  preference optimization strategy, compelling the model to focus on the product's intrinsic information to generate both visually appealing and contextually consistent advertising images.
\end{itemize}
\vspace{-0.3em}

Extensive experiments on both public and commercial datasets demonstrate that our method achieves state-of-the-art performance across multiple key metrics, significantly improving online CTRs in real-world e-commerce scenarios.

\section{Related Works}
\subsection{Advertising Image Generation}

The primary goal of advertising image generation is to create natural and contextually relevant images while preserving the integrity and identity of the original product. Initially, template-based methods~\cite{wei2022towards,chen2103automated,wei2022towards,mishra2020learning} were employed for assembling advertising images, offering high efficiency but lacking personalization and flexibility.
With the advent of generative adversarial networks (GANs)~\cite{goodfellow2020generative}, researchers began exploring more flexible and automated approaches to advertising image creation. Ku et al.~\cite{ku2023staging} introduced a novel approach of using GAN models as retrieval-assisted techniques for enhancing product images in advertising contexts.
More recently, diffusion models have shown promise in producing high-quality, realistic ad images. InsertDiffusion~\cite{mueller2024insertdiffusion} introduced a training-free diffusion architecture that effectively embeds objects into images while preserving their structural and identity features. 
Recognizing that ad quality involves multiple aspects such as aesthetics and text-image consistency, researchers have begun exploring multi-stage optimization methods~\cite{li2023planning,chen2024virtualmodel,lee2024parrot}. A notable example is VirtualModel~\cite{chen2024virtualmodel}, which employs a multi-branch structure to enhance the credibility of human-object interactions and ensure consistency in generation quality. 
Unlike previous methods primarily focusing on visual quality or text-image consistency, our method uniquely leverages MLLMs to generate CTR-optimized contextual descriptions, guiding diffusion models to produce visually appealing and product-specific advertising images.

\subsection{Click-Through Rate Prediction}

Click-Through Rate (CTR) prediction plays a crucial role in online advertising and recommendation systems, directly impacting user experience and revenue generation. In the context of CTR-driven advertising image generation, precise CTR estimation enables more effective selection and positioning of visual content, thereby enhancing the overall performance of online advertising campaigns.
The advent of deep learning has revolutionized traditional CTR prediction~\cite{kumar2015predicting,juan2016field,jie2017ctr}, enabling models to automatically learn hierarchical feature representations from raw input data.
This paradigm shift not only improved the performance of textual or numerical-based CTR prediction methods~\cite{kumar2015predicting,jie2017ctr} but also paved the way for incorporating visual elements into the prediction process.
For instance, Wang et al.~\cite{wang2021hybrid} proposed a hybrid bandit approach that integrates visual priors with a dynamic ranking mechanism, demonstrating the potential of incorporating visual information in CTR prediction models.
Recognizing that real-world advertisements are inherently multimodal, comprising text, visuals, and other data types, researchers have begun to explore methods that can effectively integrate these diverse modalities. CG4CTR~\cite{yang2024new} leveraged a multi-head self-attention module to jointly process textual and visual information from multimodal advertisements, extracting rich features for more accurate CTR estimation.
However, these approaches often struggle with complex image understanding tasks and fail to effectively integrate multimodal information. Therefore, it is imperative to explore a more robust CTR estimation method that can seamlessly interpret visual content and harmoniously fuse information from multiple modalities.

\subsection{Learning from Human Feedback}

Reinforcement Learning from Human Feedback (RLHF)~\cite{ziegler2019fine,bai2022training,ouyang2022training} involves collecting human feedback on model outputs. This feedback is then used to optimize the generation model using reinforcement learning algorithms such as PPO\cite{schulman2017proximal} or DPO~\cite{rafailov2024direct}.
For example, Lee et al.~\cite{lee2023aligning} proposed a three-stage fine-tuning method to improve text-image alignment in text-to-image (T2I) models using human feedback and reward-weighted likelihood maximization.
Wu et al.\cite{wu2023better} introduced a human preference score derived from a classifier trained on human-curated image choices, which is then utilized to adapt T2I models. Parrot~\cite{lee2024parrot} proposed a multi-reward RL approach that jointly optimizes the T2I model and prompt expansion network to improve image quality.
However, current preference optimization methods for image generation, while showing promise in text-to-image (T2I) tasks, face significant challenges when applied to scenarios with strict visual requirements, such as advertising background generation. These methods often focus solely on optimizing specific metrics, neglecting the contextual relevance and visual harmony of the generated content. 
Therefore, our method emphasizes exploring optimization techniques that enable the model to effectively integrate multimodal information to generate diverse and coherent background descriptions that better align with user preferences.

\begin{figure*}[t!]
    \centering
    \includegraphics[width=0.95\textwidth]{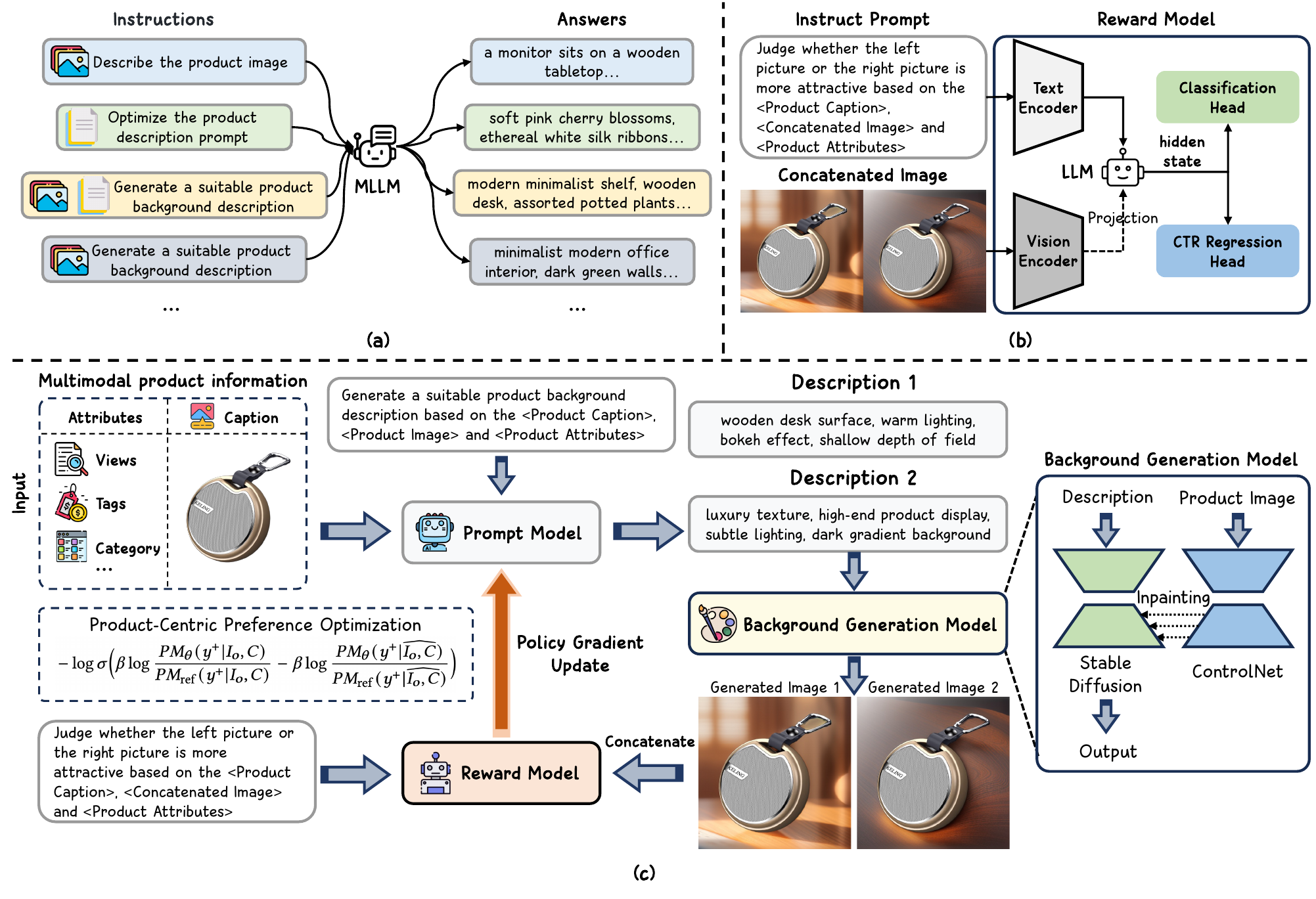}
    \vspace{-1em}
    \caption{(a) E-commerce knowledge pre-training. The MLLM is pre-trained on a large-scale multimodal e-commerce dataset to incorporate domain-specific knowledge. (b) The Structure of RM. The RM integrates multimodal product features using visual and textual encoders, with dual branches to estimate CTR and identify appealing ad images. (c)  CTR-driven preference optimization stage. The PM generates background descriptions for background generation model to create product images with various backgrounds. The RM then estimates the CTR for these images, simulating human feedback to optimize the PM.}
    \label{fig:1}
\end{figure*}

\section{Method}
\subsection{Overview}

In this work, we introduce a novel method called \textbf{C}TR-Driven \textbf{A}dvertising \textbf{I}mage \textbf{G}eneration (CAIG), designed to generate compelling advertising images that capture user interest, as shown in Figure~\ref{fig:1}.
We first pre-train the MLLM on a large-scale multimodal e-commerce dataset, injecting domain-specific knowledge into the model. This serves as the foundation for our Prompt Model (PM) and Reward Model (RM). Then, we initialize the RM from the pre-trained MLLM and further train it on extensive multimodal online user click data, enabling the RM to simulate human feedback.
Finally, we introduce a CTR-driven preference optimization stage, which adopts Product-Centric Preference Optimization (PCPO) as its core strategy, detailed in Algorithm~\ref{alg:caig}. This stage uses the RM's feedback to fine-tune the PM, ultimately generating advertising images that balance attractiveness and relevance.

\subsection{E-commerce Knowledge Pre-training}

To address the challenge of efficient and scalable advertising creative generation, we leverage the power of MLLMs by injecting domain-specific e-commerce knowledge through pre-training on a large-scale multimodal e-commerce dataset comprising 1.2M samples from major e-commerce platforms, as shown in Figure~\ref{fig:1} (a). Specifically, the pre-training tasks involve three main tasks:
\vspace{0.5em}
\begin{enumerate}
\item \textbf{Image Understanding:} Describing the products or backgrounds based on product images.
\item \textbf{Multimodal Content Comprehension:} Describing product background or generating product titles based on multimodal product information (e.g., titles, categories, tags).
\item \textbf{Prompt Generation:} Generating or rewriting description prompts based on multimodal product information.
\end{enumerate}
\vspace{0.5em}

To facilitate the model's understanding of product information, we design an instruction function that elegantly integrates diverse product attributes into a unified, semantically rich description. Formally, this can be expressed as:
\begin{gather}
C = f_{\text{instruct}}(Q,i_1,i_2,...,i_n),
\end{gather}
where $C$ is the instruct prompt constructed by the instruct function $f_{instruct}$  from $n$ individual product attributes $\textbf{i}=[i_1,i_2,...,i_n]$ (such as title, category, price, etc.), $Q$ is the task-specific question. For example, an instruct statement for a specific product might be formulated as:
\textit{"Generate a suitable product background description based on the following attributes: Product Title: 'Wireless Bluetooth Earbuds', Product Category: 'Electronics', Price: '\$49.99', Customer Rating: '4.5 stars', Color Options: 'Black, White, Blue'"}.

By leveraging the power of MLLMs and our specialized pre-training tasks, our MLLM gains a deep understanding of e-commerce products and their attributes. This understanding lays a solid foundation for vision-based CTR prediction and advertising image generation in subsequent tasks, enabling the creation of more relevant and engaging visual content for e-commerce advertising.

\begin{algorithm}[t]
\caption{CTR-Driven Preference Optimization}
\label{alg:caig}
\renewcommand{\algorithmicrequire}{\textbf{Input:}}
\renewcommand{\algorithmicensure}{\textbf{Output:}}
\begin{algorithmic}[1]
\REQUIRE
        $N$ -- Number of training epochs \\
        $M$ -- Number of products \\
        $\mathrm{PM}_\Theta$ -- Pre-trained Prompt Model \\
        $\mathrm{RM}_\Theta$ -- Pre-trained Reward Model \\
\FOR{ \textit{epoch} $i = 1$ to $N$}
    \STATE $P \leftarrow \emptyset$ Initialize set for positive and negative sample pairs
    
    \FOR{ \textit{product} $j = 1$ to $M$}
        \STATE $(I_o, C) \leftarrow$ Get product image and instruct prompt 
        \STATE $(y_{1}, y_{2}) \leftarrow \mathrm{PM}_\Theta(I_o, C)$ Generate two background descriptions

        \STATE $(I_{1}, I_{2}) \leftarrow$  Generate advertising images using Stable Diffusion and ControlNet with $I_o$ and $(y_{1}, y_{2})$
        
        \STATE $(p_1, p_2) \leftarrow \mathrm{RM}_{\Theta}([I_{1}; I_{2}], C)$ \hfill Predict relative CTR by RM 

        \STATE $(y^+, y^-) \leftarrow (y_1, y_2) \text{ if } p_1 > p_2 \text{ else } (y_2, y_1)$
        \STATE $P \leftarrow P \cup \{(I_o, C, y^+, y^-)\}$  Insert preference pair
        
        \ENDFOR
        
        \STATE Update $\mathrm{PM}_\Theta$ using $P$ with $\mathcal{L}_{\mathrm{DPO}} + \mathcal{L}_{\mathrm{PCPO}}$ (Equation 12)
        
\ENDFOR
\ENSURE 
        $\mathrm{PM}_\Theta$ -- Fine-tuned Prompt Model
\end{algorithmic}
\end{algorithm}
\subsection{Reward Model based on MLLM}

To optimize the alignment between generated advertising images and online user preferences, we leverage user feedback data to train a RM for fine-tuning the advertising image generation pipeline. First, our method utilizes the strong visual representation capabilities and flexible multimodal input of the MLLM which is pre-trained with e-commerce knowledge to extract robust product features. Furthermore, to mitigate the impact of absolute CTR variations across different product categories, we reformulate the CTR regression task into a relative  comparison task between pairs of images, as illustrated in Figure~\ref{fig:1} (b).

Specifically, we construct pair-wise training samples from user click data, where each pair consists of two advertising images for the same product with corresponding CTRs. For each product pair $(I_1,I_2)$ sharing common attributes $\mathbf{i}=[i_1,i_2,...,i_n]$, we first create an instruct prompt $C_{\text{RM}}$ by integrating the product attributes with the RM-specific question template $Q_{\text{RM}}$ through the prompt engineering function $f_{\text{instruct}}$. The multimodal input is then formed by concatenating the visual representations of both images with the textual prompt embedding, which can be formally expressed as:
\begin{gather}
    C_{\text{RM}} = f_{\text{instruct}}(Q_{\text{RM}},i_1,i_2,...,i_n), \\
    H = \mathrm{LLM}\left([f_{\mathrm{vision}}([I_1;I_2]); f_{\mathrm{text}}(C_{\text{RM}})]\right),
\end{gather}
where $f_{\mathrm{vision}}$ and $f_{\mathrm{text}}$ denote the vision and text encoders respectively. The LLM processes this combined representation to produce hidden states $H\in\mathbb{R}^{l\times d}$, with $d$ representing the hidden dimension and $l$ the sequence length. Following standard practice in sequence classification with LLMs~\cite{touvron2023llama}, we leverage the hidden state of the final token $h\in\mathbb{R}^d$ as the discriminative representation, capturing the cumulative contextual information of the complete input sequence.

Subsequently, we transform the CTR regression task into a binary classification problem that directly compares the relative CTR performance between the left and right images in each pair. A classification head $FC_{\text{cls}}$ is employed to map the final token's hidden state $h$ to a two-dimensional probability distribution $p\in\mathbb{R}^2$:
\begin{equation}
p = \text{softmax}(FC_{\text{cls}}(h)).
\end{equation}
To train the RM, we initialize it with pre-trained weights infused with e-commerce domain knowledge and utilize the binary cross-entropy loss function for training. The loss function is defined as:
\begin{equation}
\mathcal{L}_{\text{CE}} = -\sum_{i=1}^{N} [t_i^T \log(p_i)],
\end{equation}
where $N$ is the number of training samples, $t_i\in\{[1,0],[0,1]\}$  indicates whether the left or right side of the concatenated image has a higher CTR, and $p_i$ is the predicted probability distribution.

Additionally, to enable the model to predict the CTR of the left and right images in a composite image with fine-grained accuracy, we introduce a point-wise loss using a separate CTR regression branch:
\begin{equation}
\mathcal{L}_{\text{Point}} = \frac{1}{N}\sum_{i=1}^{N} ||FC_{\text{ctr}}(h_i) - \Hat{t_i})||_2^2,
\end{equation}
where $FC_{\text{ctr}}$ represents the fully connected layer for CTR regression, $FC_{\text{ctr}}(h_i)$ represents the predicted CTR values for the $i$-th image pair, and $\Hat{t_i} \in \mathbb{R}^2$  corresponds to the true CTRs for the left and right images in the pair.

The final loss function for training the RM is a combination of the binary cross-entropy loss and the PointLoss:
\begin{equation}
\mathcal{L}_{\text{reward}} = \lambda_1 \mathcal{L}_{\text{CE}} + \lambda_2 \mathcal{L}_{\text{Point}},
\end{equation}
where $\lambda_1$ and $\lambda_2$ are hyperparameters that balance the contribution of each loss component. This combined design of two components enables the model to learn the relative CTR of comparative advertising images during the training phase while incorporating absolute CTR as an auxiliary input. During the inference stage, we utilize the comparison results from the classification head as the basis for comparing CTR.

\subsection{Product-Centric Preference Optimization}
We formulate the task of generating higher CTR advertising images as a preference selection problem, encouraging the advertising generation model to choose higher attractive positive images $I^{+}$ and reject less attractive negative images $I^{-}$.
This process involves two key steps: (1) generating image pairs and comparing their CTR with RM, (2) fine-tuning the generation model based on the feedback from the RM, as illustrated in Algorithm~\ref{alg:caig}.
For advertising image generation, we utilize the background description $y$ generated by our PM as input to Stable Diffusion~\cite{rombach2022high}, along with the original product image $I_o$. We employ ControlNet~\cite{zhang2023adding} and inpainting techniques \cite{lugmayr2022repaint} to seamlessly integrate the product into the generated background. The process uses DDIM~\cite{song2020denoising} as the denoising schedule, where the latent representation $x_t$ at step $t$ is calculated as :
\begin{gather}
x_t = \sqrt{\bar{\alpha}_t}\frac{x_{t+1}-\sqrt{1-\bar{\alpha}_{t+1}}\epsilon_\theta(x_{t+1}, y)}{\sqrt{\bar{\alpha}_{t+1}}} + \sqrt{1-\bar{\alpha}_t}\epsilon_\theta(x_{t+1},y)
\end{gather}
where $\epsilon_\theta(x_{t+1},y)$ represents the noise predicted by the model~\cite{rombach2022high, zhang2023adding}, and $\bar{\alpha}$ is a set of coefficients controlling the forward noise-adding process. The latent representation $x_t$ is processed by:
\begin{gather}
x_{t}= (\boldsymbol{I}-\boldsymbol{M})\otimes x_{t} + \boldsymbol{M}\otimes x_{o},
\end{gather}
where $x_o$ is the latent of $I_o$, $\boldsymbol{I}$ represents an identity matrix, $\boldsymbol{M}$ is product mask and $\otimes$ denotes the element-wise multiplication. The final latent $x_0$ is then converted to the generated image $I_g$.

Considering that collecting real CTR feedback is time-consuming and resource-intensive, we leverage the RM to distinguish in real-time between more attractive $I^{+}$ and  less attractive $I^{-}$ generated images to fine-tune the generation pipeline. Similar to Parrot~\cite{lee2024parrot}, we empirically find that fine-tuning the background generation model has a much smaller impact on the image content compared to changing the background description. Therefore, to enhance training efficiency, we focus solely on fine-tuning the PM to choose higher attractive background descriptions $y^{+}$ and reject less attractive ones $y^{-}$. The Direct Preference Optimization (DPO)~\cite{rafailov2024direct} is then adopted as our fundamental strategy due to its simplicity and efficiency.
Specifically, given an optimization policy model $PM_\theta$ and a reference model $PM_\mathrm{ref}$,  the DPO objective is:
{
\small
\begin{equation}
    \mathcal{L}_{\mathrm{DPO}}=-\log\sigma\Big(\beta\log\frac{PM_\theta(y^{+}|I_o,C)}{PM_\mathrm{ref}(y^{+}|I_o,C)}-\beta\log\frac{PM_\theta(y^{-}|I_o,C)}{PM_\mathrm{ref}(y^{-}|I_o,C)}\Big),
\end{equation}
}%
where $(I_o,C)$ represent the original product image and corresponding instruct prompt. $\sigma$ is the sigmoid activation function, $\beta$ is a regularization parameter. During the DPO process, the reference model $PM_\mathrm{ref}$ is frozen to optimize the policy model $PM_\theta$.

It is worth noting that excessive focus on CTR optimization during DPO training may ignore the product information in preference data, causing a mismatch between the foreground and background in the generated image.
Therefore, we introduce the Product-Centric Preference Optimization (PCPO).
The core mechanism of PCPO is to control product information as the sole variable during the training process and construct additional preference data pairs, thereby encouraging the model to generate background descriptions that match the product characteristics. Specifically, given a matched $(y^{+},I_o,C)$ and a mismatched $(y^{+},\widehat{I_o,C})$, the PCPO objective is formulated as:
{
\small
\begin{equation}
    \mathcal{L}_{\mathrm{PCPO}}=-\log\sigma\Big(\beta\log\frac{PM_\theta(y^{+}|I_o,C)}{PM_\mathrm{ref}(y^{+}|I_o,C)}-\beta\log\frac{PM_\theta(y^{+}|\widehat{I_o,C})}{PM_\mathrm{ref}(y^{+}|\widehat{I_o,C})}\Big).
\end{equation}
}%

We consider two strategies to construct a product information $(\widehat{I_o,C})$ that mismatches $y^{+}$: (1) visual-aware optimization: randomly masking 75\% of input product images. (2) textual-aware optimization: randomly selecting and replacing textual information from other products. These strategies are designed to create hard negative samples that are not fully compatible with $y^{+}$ while retaining some common features with the original input.
The total objective is a combination of the standard DPO and PCPO:
\begin{equation}
\mathcal{L}_{\mathrm{opt}}=\mathcal{L}_{\mathrm{DPO}}+\mathcal{L}_{\mathrm{PCPO}}.
\end{equation}

Finally, we utilize the fine-tuned PM to generate background descriptions for products. These descriptions are then fed into the background generation model to create product advertising images suitable for online environments.

\begin{figure*}[t]
    \centering
    \begin{subfigure}[b]{0.48\textwidth}
        \centering
        \includegraphics[width=\textwidth]{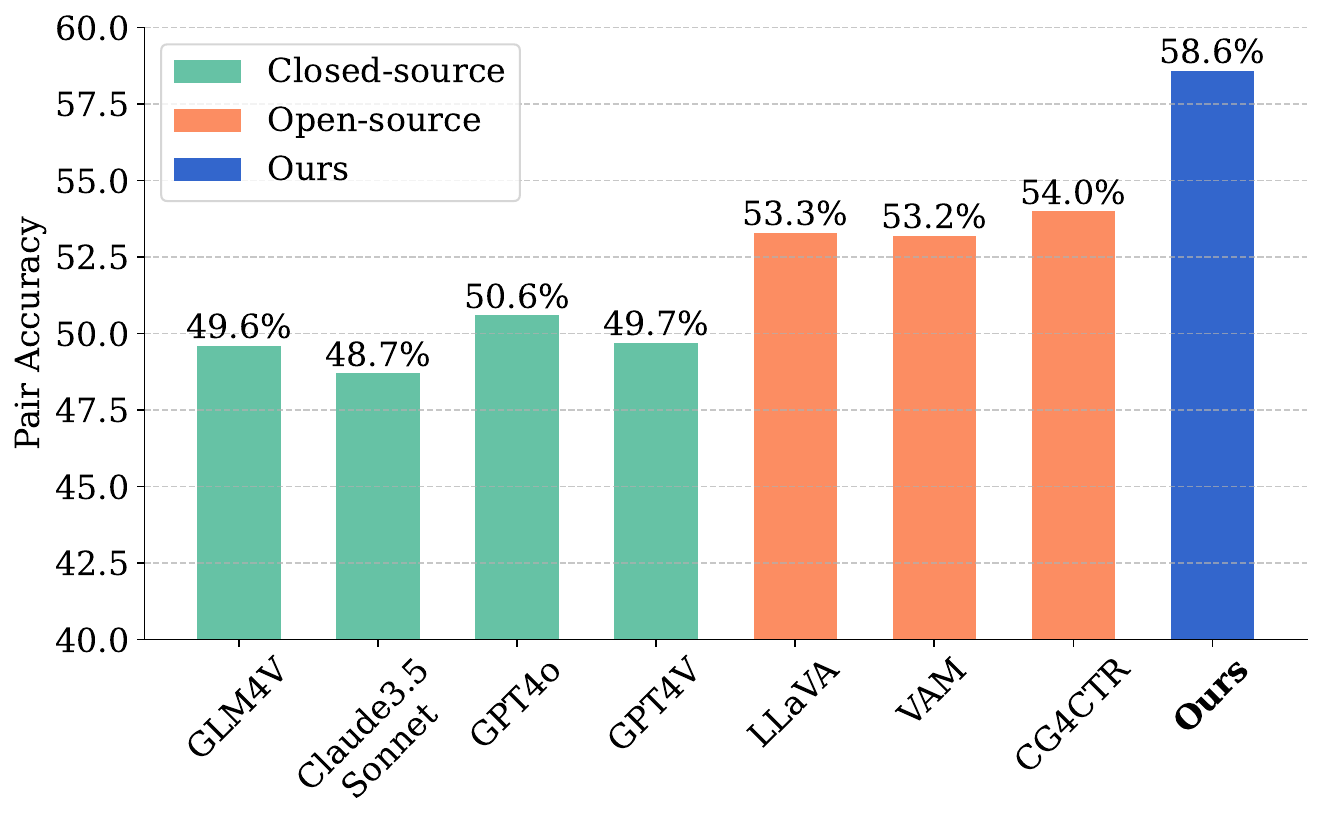}
        \vspace{-2.3em}
        \caption{Commercial data}
        \label{fig:first}
    \end{subfigure}
    \hfill
    \begin{subfigure}[b]{0.48\textwidth}
        \centering
        \includegraphics[width=\textwidth]{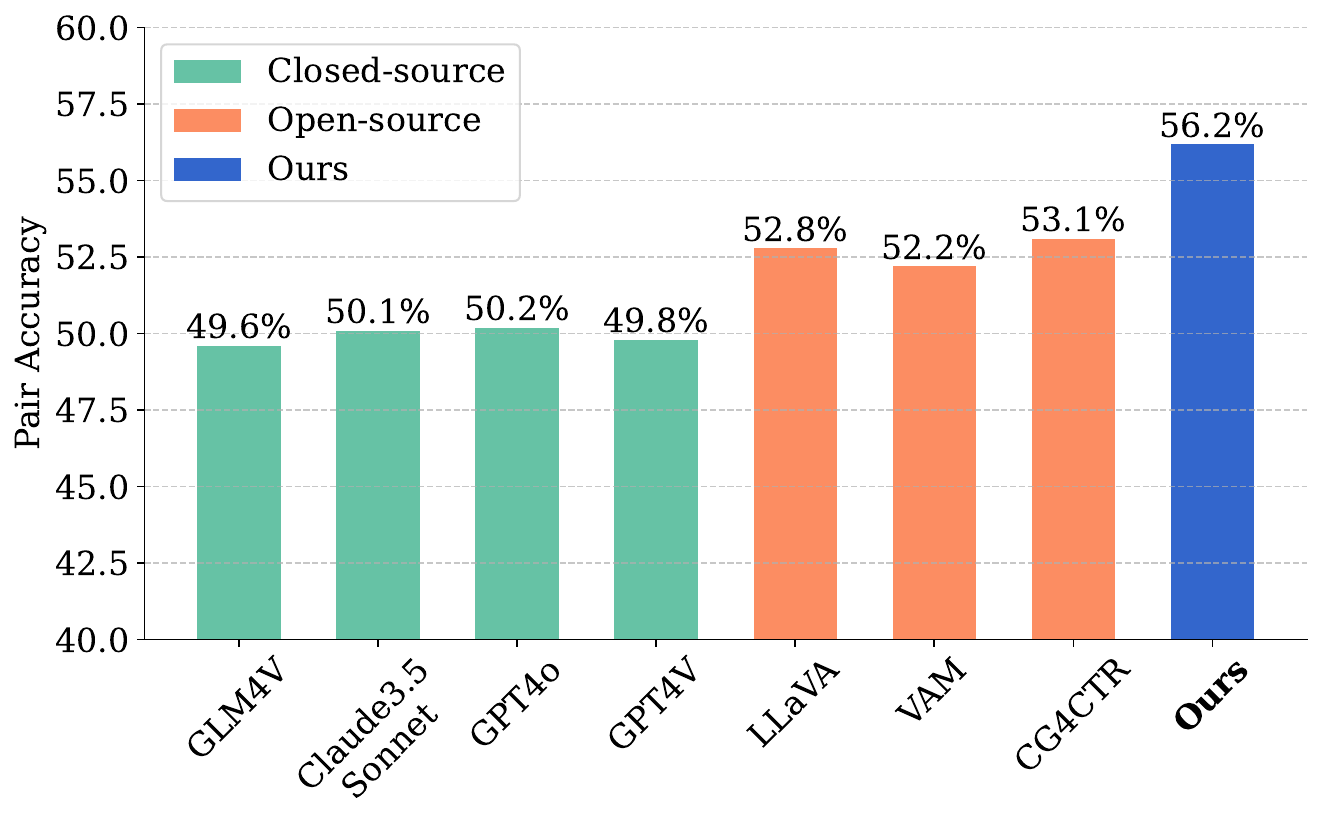}
        \vspace{-2.3em}
        \caption{Public data~\cite{wang2021hybrid}}
        \label{fig:second}
    \end{subfigure}
\vspace{-1em}
\caption{Comparison of Pair Accuracy across different methods on commercial and public datasets.}
    \label{fig:overall}
\vspace{-0.5em}
\end{figure*}

\section{Experiments}


\subsection{Experimental Setup}

\textbf{Datasets:}
For training and validating our RM, we conduct experiments on both public and commercial datasets.
The public dataset~\cite{wang2021hybrid} covers 500K product samples with 1.2M unique advertising images. The CTR data for public dataset is collected over an average period of 10 days for each creative placement. Our commercial dataset, collected from a well-known e-commerce platform, contains 1M product samples with 3.4M unique advertising images. For commercial dataset, the CTR data is collected over a one-month period. It is worth noting that our commercial dataset contains more detailed product information, including titles, categories, tags, and other relevant attributes.
To ensure the quality and reliability of our training and test data, we apply specific criteria to both datasets. To improve the confidence of CTR estimates, we require each image to have a minimum exposure threshold (E). Additionally, to ensure distinguishable CTR differences within pairs, we require the relative CTR difference between paired images to exceed a certain threshold (D). For the training set, we set E = 50 and D = 1\%, while for the test set, we apply more stringent criteria with E = 1,000 and D = 5\%. After applying these preprocessing steps, the public dataset yields 890K training pairs and 1,034 test pairs, while our commercial dataset contains 1.15M training pairs and 1,528 test pairs.
In the  CTR-driven preference optimization stage, we fine-tune our generation pipeline using a dataset of 32K samples, which includes original product images along with multimodal information. These product samples are also collected from the same e-commerce platform, ensuring consistency in data source and characteristics.

\noindent\textbf{Models:} We employ the \textit{LLaVA-v1.6-7B}~\cite{liu2024visual} as our foundation MLLMs, which utilizes \textit{Vicuna-7B}~\cite{vicuna2023} as the text encoder and \textit{CLIP-ViT-L/14-336}~\cite{radford2021learning} as the vision encoder. For our background generation model, we use \textit{MajicmixRealistic-v7}\footnote{\url{https://civitai.com/models/43331/majicmix-realistic}} as the base model,  enhanced with \textit{ControlNet-v1.1}~\cite{zhang2023adding}\footnote{\url{https://github.com/lllyasviel/ControlNet}} to provide finer control over the generated images.

\noindent\textbf{Implementation Details:}
For the pre-training task of MLLMs, we perform full model fine-tuning. We optimize the learning process over 10 epochs using a cosine learning rate scheduler with an initial learning rate of 2e-6. The pre-training task takes approximately 5 days to complete. We then initialize the RM with the pre-trained weights and employ the same learning strategy to train on massive user click data, simulating user feedback. The hyperparameters $\lambda_1$ and $\lambda_2$ are set to 1 and 0.5, respectively. Finally, in the  CTR-driven preference optimization stage, we utilize the frozen RM to drive the proposed advertising image generation pipeline. We employ LoRA~\cite{hu2021lora} fine-tuning with a learning rate of 2e-5. This phase consists of 5 epochs and takes about 20 hours to complete. All experiments are conducted on a machine equipped with 8 NVIDIA A100 GPUs.

\subsection{Analysis on Reward Model}
\subsubsection{Evaluation Metric}
To evaluate the performance of our RM, we introduce the Pair Accuracy metric, defined as:
\begin{equation}
\text{Pair Accuracy} = \frac{1}{N}\sum_{i=1}^{N} \boldsymbol{1}(\operatorname{argmax}(p_i) == y_i),
\end{equation}
where $N$ is the total number of image pairs, $p_i$ is the predicted probability distribution for the i-th pair, $y_i$ is the ground truth label, and $\boldsymbol{1}$ is the indicator function. It is worth noting that the task of CTR comparison is highly challenging, where even small improvements can lead to significant economic benefits.

\subsubsection{Comparison with State-of-the-Art Methods}

We conduct extensive experiments on both commercial and public datasets, comparing our method with various state-of-the-art open-source and closed-source models based on MLLMs, as shown in Figure~\ref{fig:overall}. The open-source models are fine-tuned on the corresponding datasets to ensure a fair comparison. For closed-source models, we provide them with the same instructions and image pairs as our RM, then convert their textual responses into predicted labels.
From the results, we can observe that existing closed-source models (GLM4V~\cite{glm2024chatglm}, Claude3.5 Sonnet~\cite{2024Claude3}, GPT4o~\cite{achiam2023gpt}, and GPT4V~\cite{2023GPT4VisionSC}) lack the ability to effectively compare the CTR of advertising images, as evidenced by their near-random performance (around 50\% Pair Accuracy). This suggests that these models, despite their general capabilities, are not specifically tuned for CTR regression tasks in advertising contexts.
Open-source models like VAM~\cite{wang2021hybrid} and CG4CTR~\cite{yang2024new}, while showing slight improvements, still demonstrate limited performance due to their weak visual representation capabilities and inability to effectively integrate multimodal information.
In contrast, our proposed method, which leverages MLLM, achieves state-of-the-art performance on both commercial and public datasets. By effectively combining visual and textual modalities of product information, our method demonstrates superior ability in predicting relative CTR performance between image pairs. Specifically, our method achieves a Pair Accuracy of 58.6\% on commercial data and 56.2\% on public data, significantly outperforming all baseline models.

\subsubsection{Ablation Study}

To further analyze the contribution of each component in our proposed RM, we conduct a detailed ablation study on the commercial dataset, with results shown in Table \ref{tab:ablation_reward_model}. We start with the base \textit{LLaVA-v1.6-Vicuna-7B}~\cite{liu2024visual} and progressively add key components to observe their impact on model performance.
First, we observe that incorporating a pre-training step with e-commerce domain knowledge increases the Pair Accuracy from 53.3\% to 54.4\%. This suggests that domain-specific pre-training provides a good starting point for the model, enhancing its baseline understanding of e-commerce concepts and product characteristics.
A more substantial improvement is seen when replacing the original output layer with a dedicated classification head, which boosts the Pair Accuracy to 56.4\%. This notable increase can be attributed to the classification head's ability to enable the model to learn explicit classification boundaries, thereby reducing the ambiguity often associated with natural language outputs in the original model architecture.
Incorporating product captions and additional product information further enhances the model's accuracy to 58.2\%. This demonstrates the importance of supplementary product data in improving the model's ability to compare advertising image attractiveness.
Finally, we add an extra CTR regression branch and introduce the point loss, which improve the final performance to 58.6\%. This enhancement demonstrates that by directly incorporating CTR values into the training objective, the model can more accurately capture subtle differences in CTR, thereby further improving prediction accuracy. 

\begin{table}[t]
\centering
\resizebox{\linewidth}{!}{
\begin{tabular}{ccccc|c}
\toprule
\begin{tabular}[c]{@{}c@{}}E-commerce\\ pre-training\end{tabular} & \begin{tabular}[c]{@{}c@{}}Classification\\ head\end{tabular} & \begin{tabular}[c]{@{}c@{}}Product\\ caption\end{tabular} & \begin{tabular}[c]{@{}c@{}}Additional\\ information\end{tabular} & Pointloss & Pair Accuracy (\%) \\
\midrule
\xmark & \xmark & \xmark & \xmark & \xmark & 53.3 \\
\cmark & \xmark & \xmark & \xmark & \xmark & 54.4 (\textcolor{blue}{+1.1\%}) \\
\cmark &\cmark & \xmark & \xmark & \xmark & 56.4 (\textcolor{blue}{+2.0\%}) \\
\cmark &\cmark & \cmark & \xmark & \xmark & 57.3 (\textcolor{blue}{+0.9\%}) \\
\cmark &\cmark & \cmark & \cmark & \xmark & 58.2 (\textcolor{blue}{+0.9\%}) \\
\cmark & \cmark & \cmark & \cmark & \cmark & 58.6 (\textcolor{blue}{+0.4\%}) \\
\bottomrule
\end{tabular}
}
\vspace{0.3em}
\caption{Ablation study for the reward model.}
\label{tab:ablation_reward_model}
\vspace{-3em}
\end{table}

\subsection{Analysis of Product-Background Matching}

\subsubsection{Evaluation Metric}

Existing preference optimization methods focus solely on optimizing rewards, which may neglect the crucial balance between visual appeal and contextual appropriateness.
To quantify the impact of different optimization methods on the compatibility between foreground products and generated backgrounds, we introduce the match rate metric. We calculate the match rate by randomly selecting 1,000 products and generating backgrounds for each using the generation models under evaluation. Experienced advertising professionals then assess whether the foreground and background are compatible based on comprehensive product information, considering factors such as style consistency, color harmony, and contextual appropriateness. Detailed annotation guidelines and criteria are provided in Appendix~\ref{sec:ann}.

\subsubsection{Comparison with Standard DPO}
\begin{figure*}[t]
    \centering
    \includegraphics[width=0.97\textwidth]{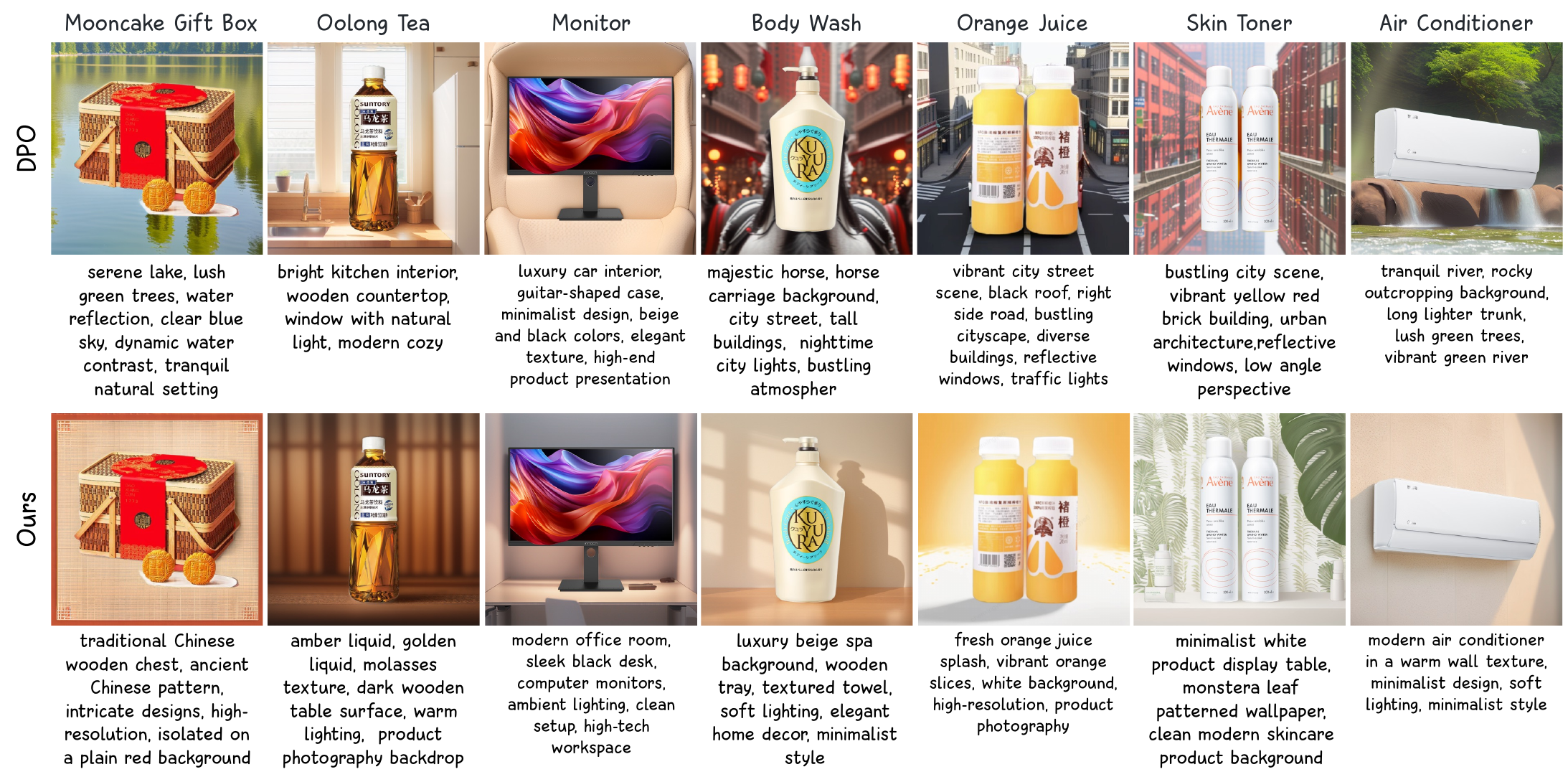}
    \vspace{-0.5em}
    \caption{Comparison between DPO and the proposed PCPO. The first line shows the name of the product, followed by the generated results for each method, including the generated image and corresponding background prompt.}
    \label{fig:mismatch}
\end{figure*}

To ensure a fair comparison, we evaluate PCPO against standard DPO during preference optimization, using identical RM for CTR feedback and equal training epochs. Figure~\ref{fig:Match_rate} illustrates the performance of both methods over training epochs. Notably, the standard DPO experiences a significant drop in match rate, declining from 0.842 to 0.597 after 5 epochs of training.
In contrast, our PCPO demonstrates a more gradual decline in match rate, maintaining a higher value of 0.798 at the 5th epoch, which represents a 33.7\% relative improvement over DPO at the same stage of training.
Additionally, we showcase several examples in Figure~\ref{fig:mismatch} where the standard DPO produces images with mismatched foreground and background elements, highlighting the effectiveness of our PCPO in preserving product-context coherence throughout the optimization process.

\subsubsection{Effectiveness of E-commerce Knowledge Pre-training}

As shown in Figure~\ref{fig:Match_rate}, we compare the performance of our pre-trained model with the original LLaVA model without pre-training (indicated by asterisks). The results demonstrate a significant improvement in match rate after injecting e-commerce knowledge through pre-training, with our model achieving a score of 0.842 compared to LLaVA's 0.753. This performance gap highlights that our pre-training strategy provides a strong initialization point for the subsequent preference optimization process, underscoring the importance of domain-specific knowledge in MLLMs.

\subsubsection{Ablation Studies}

\begin{figure}[t]
\centering
\includegraphics[width=\linewidth]{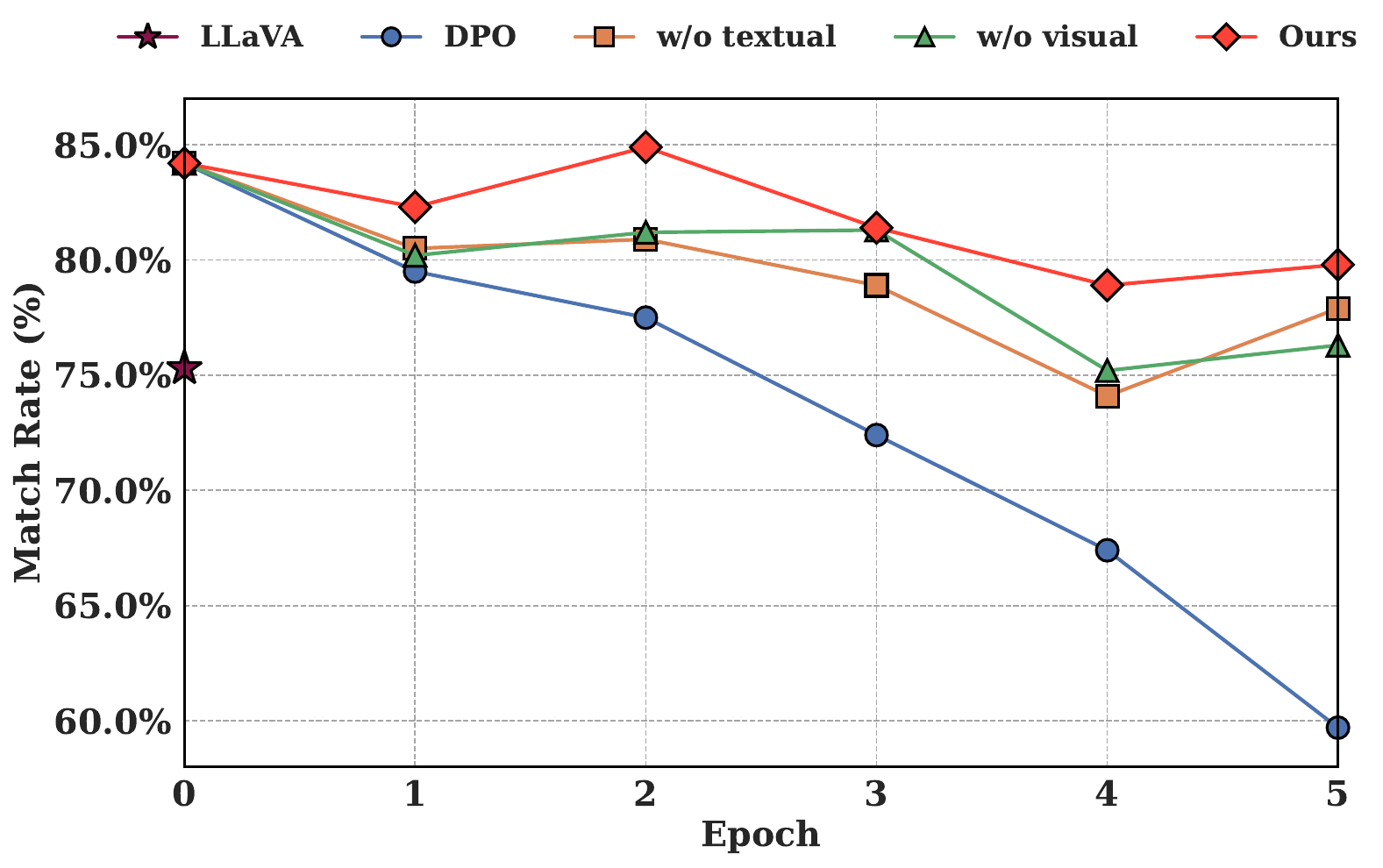}
\vspace{-2em}
\caption{Comparison of Match Rate across different preference optimization strategies over training epochs.}
\label{fig:Match_rate}
\end{figure}

To further validate the effectiveness of our method, we conduct ablation studies on two key components of PCPO: PCPO without textual-aware optimization (w/o textual) and PCPO without visual-aware optimization (w/o visual), as illustrated in Figure~\ref{fig:Match_rate}. Both ablation variants show improvements over standard DPO but fall short of the full PCPO method. The "w/o textual" and "w/o visual" variants highlight the importance of both textual and visual components in our method.
These results emphasize that controlling the textual or visual modality input as the sole variable and constructing less relevant product information as negative samples during training can effectively prevent the model from generating contextually mismatched advertisement images.
This strategy enhances the model's focus on the multimodal information of the products themselves, leading to more accurate and relevant product descriptions. The full PCPO strategy, which combines both textual and visual perturbations, is the most effective in optimizing the model's performance for product-centric tasks.


\subsection{Online Results}
\begin{table}[t]
\centering
\resizebox{\linewidth}{!}{
\begin{tabular}{l|cccccc}
\toprule
\multirow{2}{*}{Methods} & \multirow{2}{*}{All} & \multirow{2}{*}{Beauty} & \multirow{2}{*}{Fashion} & Home & \multirow{2}{*}{Digital} & \multirow{2}{*}{Computers} \\
 & & & & Appliances & & \\
\midrule
VAM~\cite{wang2021hybrid} & 3.9 & 5.1 & 3.8 & 4.2 & 2.3 & 4.5 \\
CG4CTR~\cite{yang2024new} & 3.7 & 4.8 & 3.2 & 4.9 & 2.1 & 3.7\\
DPO~\cite{rafailov2024direct} & 5.7 & 3.7 & 3.2 & 6.2 & \textbf{5.8} & 4.7\\
Ours & \textbf{7.4} & \textbf{9.5} & \textbf{7.6} & \textbf{7.8} & 3.2 & \textbf{5.4} \\
\bottomrule
\end{tabular}
}
\vspace{0.3em}
\caption{Online CTR improvement compared with the baseline of using the pre-trained MLLM with percentage.}
\label{tab:online_results}
\vspace{-1.5em}
\end{table}

To validate the effectiveness of our proposed CAIG in enhancing the CTR of generated advertising images, we conduct a one-week online experiment in a well-known e-commerce platform. We use different methods to generate two images for each product in 44 categories, which almost cover all common products, greatly exceeding the previous method~\cite{yang2024new} scope of only five categories.
It is worth noting that to enhance user experience, we engage professional advertising practitioners to ensure that the images displayed online are front and background-matched. This experiment accumulates over 10 million impressions to validate the reliability and statistical significance of the CTR results. We use a multi-armed bandit based model as online display strategy.

We report the results of different methods in all categories and five common categories in Table \ref{tab:online_results}, where the improvement of CTR is compared to directly using pre-trained MLLM. To demonstrate the superiority of our RM, we use different RMs during the CTR-drive preference optimization phase. Our RM outperforms previous methods~\cite{wang2021hybrid,yang2024new} in all categories and five common categories, demonstrating that more accurate CTR prediction can drive the generative model to produce images with higher CTR. We also compare using only DPO~\cite{rafailov2024direct} as the optimization algorithm, and the results show that using our PCPO can enable the generated model to focus on product characteristics, resulting in an increase in CTR. We further conduct an online A/B test to verify the attractiveness of our generated images, and the results show that adding these images improves 2\% in CTR with over 60 million impressions.

\section{Conclusion}

In this paper, we present an innovative \textbf{C}TR-Driven \textbf{A}dvertising \textbf{I}mage \textbf{G}eneration (CAIG) method, leveraging the powerful capabilities of Multimodal Large Language Models (MLLMs) to successfully address the limitations in optimizing online performance metrics. Our comprehensive framework, comprising targeted pre-training tasks, an MLLM-based two-branch reward model, and a product-centric preference optimization strategy, enables the generation of visually appealing and product-relevant advertising images. Extensive experiments demonstrate that CAIG achieves state-of-the-art performance in both online and offline metrics, significantly improving CTR in real-world e-commerce scenarios. This work not only advances the field of advertising image generation but also opens up new possibilities for applying MLLMs to complex multimodal tasks in e-commerce and digital advertising, laying a solid foundation for future research in this domain.

\begin{acks}
This work was partially supported by the National Key R\&D Program of China 2022YFC3301000 and Knowledge Innovation Program of Wuhan-Shuguang Project under Grant 2023010201020226.
\end{acks}

\bibliographystyle{ACM-Reference-Format}
\bibliography{sample-base}

\appendix
\section{Appendices}
This supplementary material provides:
\begin{enumerate}
\item \textbf{Section~\ref{sec:pretrain_vis}.} Visualization and analysis of the multi-task pre-training effectiveness.
\item \textbf{Section~\ref{sec:ann}.} Detailed annotation guidelines and criteria used for evaluating generated images.
\item \textbf{Section~\ref{sec:more_visual}.} Extensive  visual examples demonstrating the capabilities of CAIG across various product categories.
\item \textbf{Section~\ref{sec:instruction}.} The composition of pre-training tasks and design of instruction sets.
\item \textbf{Section~\ref{sec:Limitations}.} Discussion on current limitations of the method and proposed directions for future research.
\item \textbf{Section~\ref{sec:social}.} Analysis of potential social impacts, ethical considerations, and safeguards implemented.
\end{enumerate}

\subsection{Visualization of Pre-training Model}
\label{sec:pretrain_vis}

\begin{figure*}[t]
    \centering
    \includegraphics[width=\linewidth]{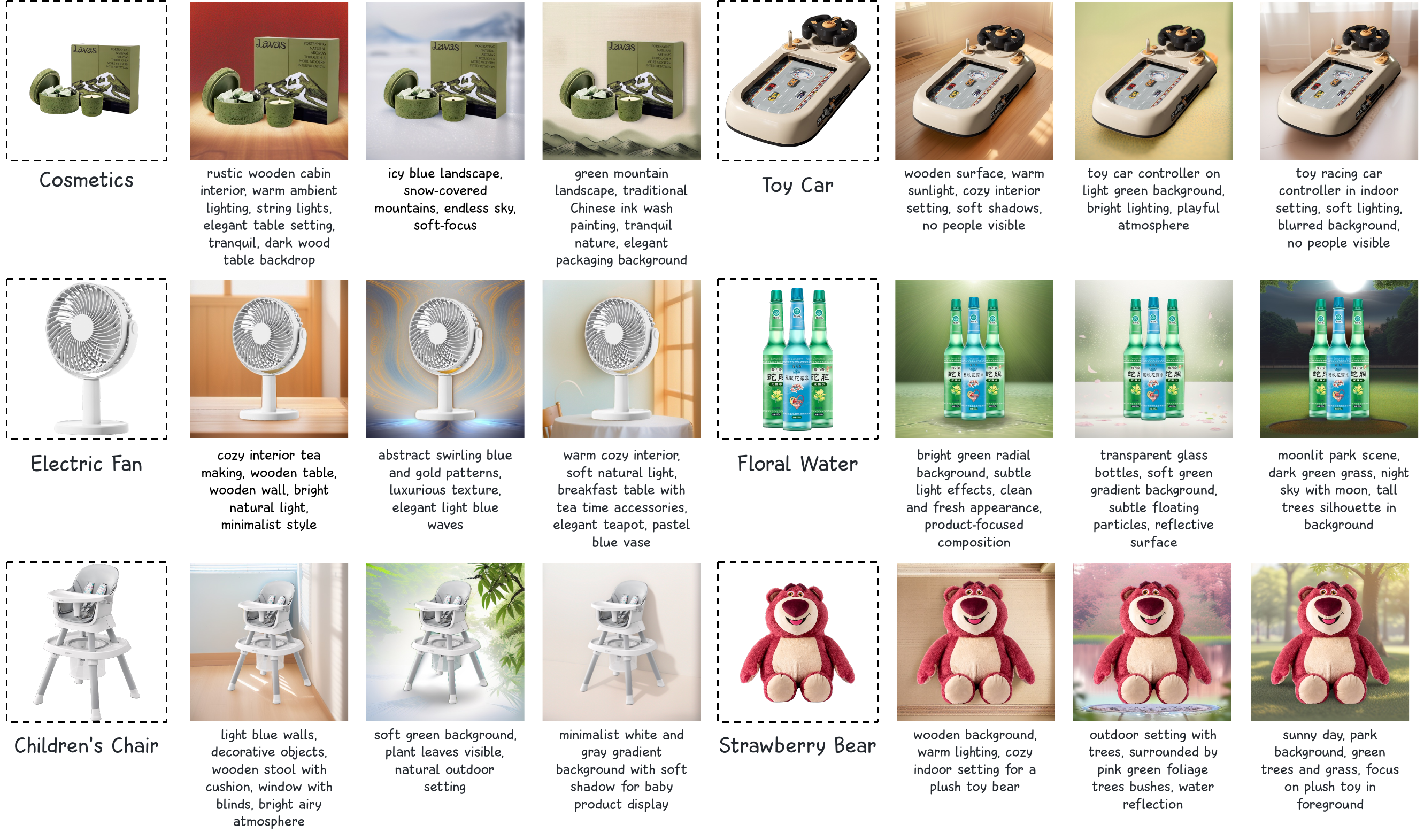}
    \vspace{-2em}
    \caption{Advertising images generated by directly using the e-commerce knowledge-injected MLLM as PM. For each product, we display the original transparent background product image in the first column, along with three different background images generated through random repetition.}
    \label{fig:pretrain_vis}
\end{figure*}

To validate the effectiveness of our proposed e-commerce knowledge injection pre-training method, we directly utilize the model pre-trained at this stage as a prompt model to generate a set of product images, as illustrated in Figure~\ref{fig:pretrain_vis}.
The generated images demonstrate the model's ability to capture key visual attributes and styles commonly found in e-commerce product photography, such as visual focus with the product as the main subject. This visual evidence suggests that our pre-training method successfully incorporated domain-specific knowledge, resulting in a model capable of generating contextually relevant and visually coherent product images.
Furthermore, the quality and diversity of the generated images indicate that the pre-trained MLLM provides a well-initialized distribution space for the subsequent preference optimization phase based on the CTR objective.

\subsection{Annotation Guidance}
\label{sec:ann}
During the match rate evaluation stage of the generation process, annotators are provided with the original product image, product title, and the generated image, along with the following strict guidelines regarding mismatches:
\begin{enumerate}
\item \textbf{Scale Mismatch.} Images where the relative size of the product and background elements are disproportionate, such as a washing machine next to an oversized laundry detergent bottle.
\item \textbf{Scene Mismatch.} Images where the product is placed in a setting that contradicts its intended use or cultural context, such as winter coats displayed in a tropical beach scene.
\item \textbf{Color Mismatch.} Images exhibiting stark color conflicts between the product and background, creating visual discomfort or detracting from the product's appeal.
\item \textbf{Available.} Images deemed suitable for advertising purposes, not falling into any of the aforementioned categories.
\end{enumerate}

\begin{figure*}[h]
    \centering
    \includegraphics[width=\linewidth]{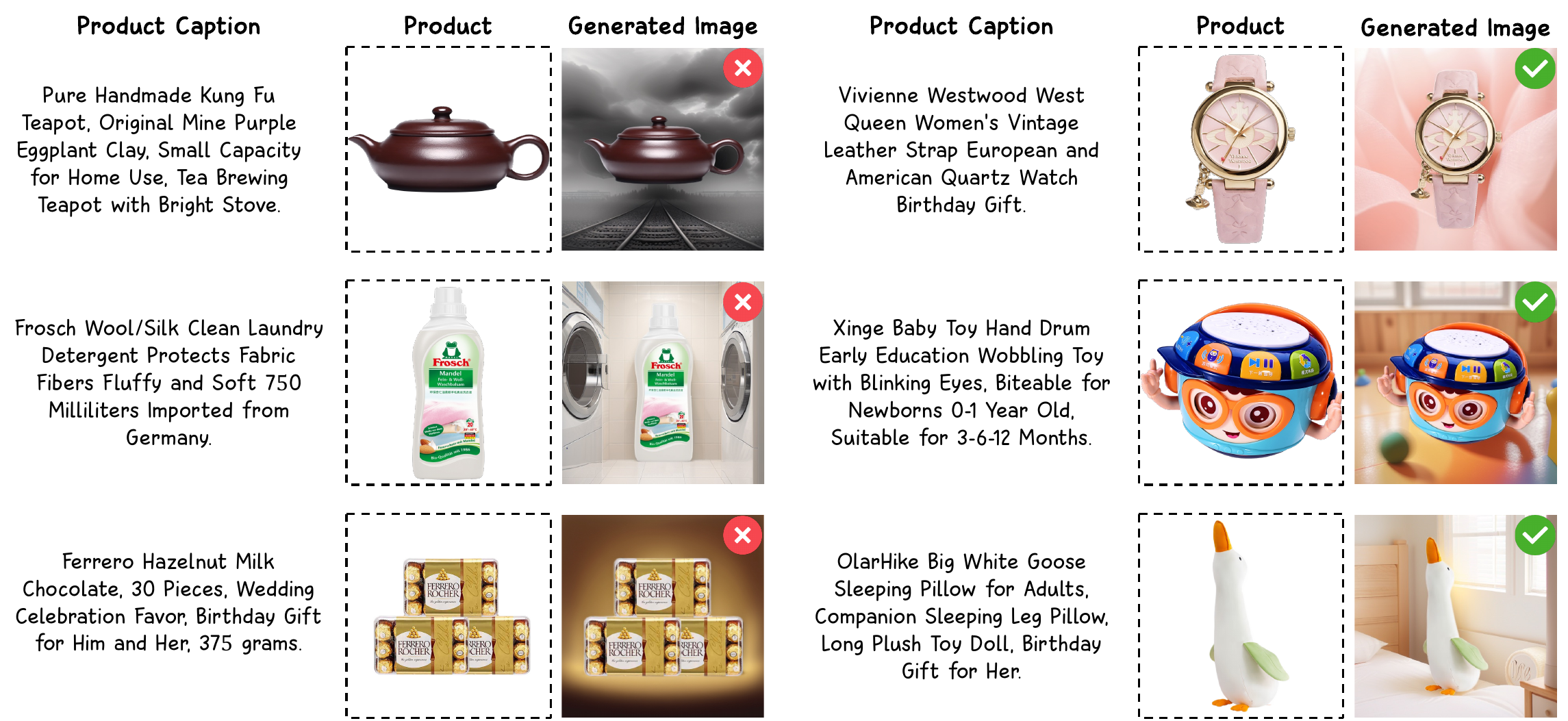}
    \vspace{-1em}
    \caption{Some match and mismatch examples identified by annotators.}
    \label{fig:guide}
\end{figure*}

Additionally, Figure~\ref{fig:guide} illustrates some examples identified by the annotators.

\subsection{More Visual Examples}
\label{sec:more_visual}

\begin{figure*}[t]
    \centering
    \includegraphics[width=\linewidth]{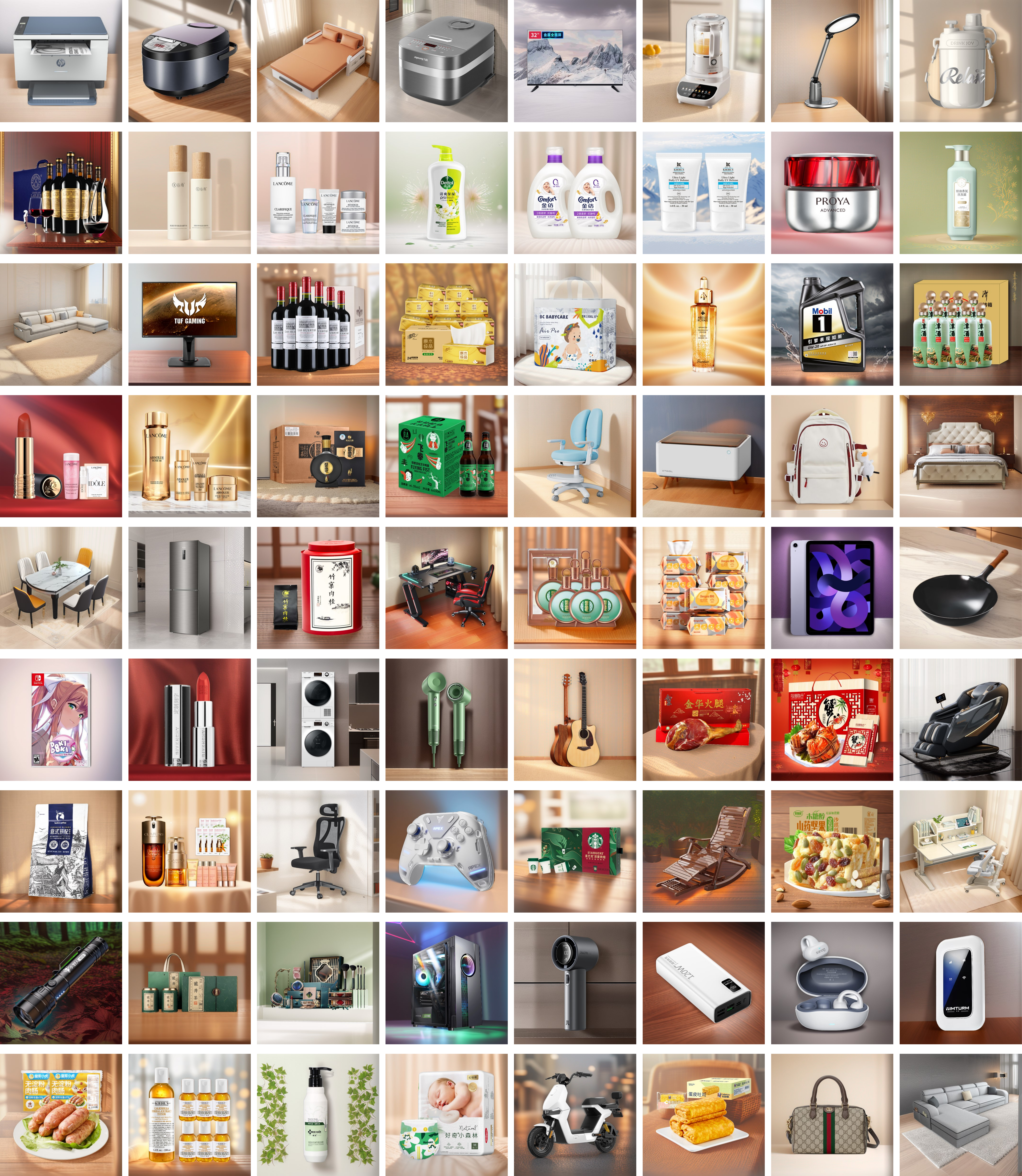}
    \caption{Extensive visual examples of our CAIG method applied to diverse product categories.}
    \label{fig:more_vis}
\end{figure*}

As illustrated in Figure~\ref{fig:more_vis}, we present an extensive array of additional examples showcasing our proposed CAIG method. These diverse visual results demonstrate the remarkable versatility and effectiveness of our method across a wide spectrum of product categories. From electronics to fashion items, and from household goods to specialty products, our method consistently generates varied and contextually appropriate backgrounds. This comprehensive set of examples not only highlights the robustness of CAIG in handling diverse product types but also underscores its ability to create visually appealing and relevant contextual environments.

\vspace{-0.5em}
\subsection{Pre-training Tasks and Instruction Set}
\label{sec:instruction}

Our pre-training method for high-quality MLLMs in advertising background generation encompasses diverse tasks and instruction sets, as illustrated in Table~\ref{tab:mulit_tasks}. We utilize a mix of public and proprietary datasets: <Product Images> and <Product Caption> from our e-commerce knowledge pre-training dataset, <Prompt> from both Promptist~\cite{hao2024optimizing} and our dataset, and COCO Caption~\cite{lin2014microsoft} for unconstrained background description generation. All target outputs are generated by GPT4V~\cite{2023GPT4VisionSC} and subsequently reviewed by experienced annotators to ensure quality and relevance.
Additionally, we design diverse instruction sets for the PM and RM, as shown in Table~\ref{tab:instruction_sets}. These guide the models in generating and evaluating advertising backgrounds from various perspectives, leveraging multimodal product information. The PM set contains 8 distinct prompts for background creation, while the RM set includes 13 distinct prompts for the CTR comparison task.
\vspace{-0.5em}
\subsection{Limitations and Future Work}
\label{sec:Limitations}

A key limitation of this work is that our CTR optimization is based on aggregated data from all users, which may overlook the preferences of minority user groups or niche market segments. This lack of personalization could result in suboptimal experiences for diverse user segments. In future work, we plan to explore personalized RLHF~\cite{siththaranjan2023distributional, zhou2023beyond, poddar2024personalizing} to better capture and integrate individual user preferences. By doing so, we aim to develop more inclusive and tailored advertising strategies that cater to a wider range of user needs and behaviors.
\vspace{-0.5em}
\subsection{Social Impact}
\label{sec:social}

Regarding image processing and automatic advertisement image generation, there are risks of producing unethical or illegal content, such as infringing on personal portrait rights or creating discriminatory content. Therefore, these technologies require stricter regulation. During the generation process, we use Stable Diffusion's official safety checker to filter out inappropriate content. We also ensure that the generated images do not contain portraits or other elements that may infringe on privacy. Finally, professionals review and screen the generated images to ensure they are free from bias or offensive content and comply with relevant laws.
To ensure the ethical use of AI in advertising, we maintain transparency by clearly labeling AI-generated images and adhering to established commercial and ethical guidelines. The automation of creative tasks may alter the job market in the creative industry. We should view AI as an auxiliary tool for creative professionals rather than a replacement to maintain the important role of human creativity in advertising.

\begin{table*}[h]
\centering
\begin{minipage}[t]{\textwidth} 
    \centering
    \begin{tabularx}{\textwidth}{X|c|c|c}
        \toprule
        \textbf{Pre-training Task} & \textbf{Input} & \textbf{Target} & \textbf{Nums} \\
        \midrule
        \multirow{2}{*}{Image Understanding\(^{3}\)} & <Product Images> & <Image Description> & 5w \\
         & <Product Images> & <Product Background Description> & 5w \\
         \midrule
         \multirow{2}{*}{Multimodal Content Understanding\(^{3}\)} & <Product Images><Product Caption> & <Image Description> & 5w \\
         & <Product Images><Product Caption> & <Product Background Description> & 15w \\
         \midrule
        \multirow{5}{*}{Prompt Generation\(^{1,2,3}\)}& <None> &  <Prompt> & 5w \\
        & <Prompt> & <Optimized prompt> & 5w \\
         & <Product Images> & <Prompt> & 10w \\
         & <Product Caption> & <Prompt> & 10w \\
         & <Product Images><Product Caption> & <Prompt> & 61w \\
         \midrule
         \multicolumn{3}{c|}{\textbf{Total}} & 121w \\
        \bottomrule
    \end{tabularx}
    \caption{Overview of pre-training tasks, input-target pairs, and data volume for our multimodal model. The tasks include image understanding, multimodal content understanding, and prompt generation, utilizing both public datasets COCO Caption\(^1\)~\cite{lin2014microsoft}, Promptist\(^2\)~\cite{hao2024optimizing} and our e-commerce knowledge pre-training dataset\(^3\).}
    \label{tab:mulit_tasks}
\end{minipage}

\vspace{1em} 

\begin{minipage}[t]{\textwidth}
    \centering
    \begin{tabularx}{\textwidth}{l|X}
    \toprule
    \multicolumn{1}{c|}{\textbf{Model}} & \multicolumn{1}{c}{\textbf{Instructions}} \\
    \midrule
    \multirow{18}{*}{Prompt Model} 
    & \textit{Design a concise Stable Diffusion prompt that takes the product caption '\{\}' and product image as inspiration to generate an appealing advertising image background for this product.} \\[3ex]
    & \textit{Produce a short diffusion prompt considering the critical information in product caption '\{\}' and product image to create an advertisement background.} \\[3ex]
    & \textit{Generate a short Stable Diffusion prompt that leverage the product caption '\{\}' and the visual elements of the product image to output an advertising background that underscores the product's attractiveness.} \\[3ex]
    & \textit{Provide a succinct text to background diffusion model prompt suitable for this product according to caption '\{\}' and product image.} \\[3ex]
    & \textit{Develop a compact prompt for Stable Diffusion to craft a background for an ad, using the product caption '\{\}' and the accompanying product image as creative influences.} \\[3ex]
    & \textit{Draft a distilled text to image prompt to fabricate an ad background, infusing the essence of '\{\}' from the product caption and the picture to accentuate the product's features.} \\[3ex]
    & \textit{Based on this product image and product caption '\{\}', formulate a brief diffusion prompt to synthesize a background tailored for advertising purposes.} \\[3ex]
    & \textit{Make use of product title '\{\}' along with its image, assemble a terse stable diffusion model prompt to render an ad background that complements and highlights the product.} \\
    \midrule
    \multirow{14}{*}{Reward Model} 
    & \textit{Comparing the left part and right part of this image, which part is more suitable for the product '\{\}'?} \\[1ex]
    & \textit{The left and right part of this image is one advertising image for the product '\{\}', respectively, which is preferred by the user?} \\[1ex]
    & \textit{Which part will bring more click-through rate in this image for product '\{\}'?} \\[1ex]
    & \textit{Between the left and right sections of this image, which one is more appropriate for showcasing the product '\{\}'?} \\[1ex]
    & \textit{Considering the left and right halves of this image, which side better represents the product '\{\}'?} \\[1ex]
    & \textit{Which side of this image, left or right, is more effective for advertising the product '\{\}'?} \\[1ex]
    & \textit{For the product '\{\}', which part of the image, left or right, is more appealing?} \\[1ex]
    & \textit{When looking at the left and right portions of this image, which part is more suitable for promoting the product '\{\}'?} \\[1ex]
    & \textit{In this image, which side, left or right, is preferred by users for the product '\{\}'?} \\[1ex]
    & \textit{Which half of this image, left or right, is more likely to attract user preference for the product '\{\}'?} \\[1ex]
    & \textit{Which section of this image, left or right, is expected to generate a higher click-through rate for the product '\{\}'?} \\[1ex]
    & \textit{For the product '\{\}', which side of the image do users find more engaging, left or right?} \\[1ex]
    & \textit{Between the left and right sides of this image, which one is anticipated to drive more clicks for the product '\{\}'?} \\
    \bottomrule
    \end{tabularx}
    \caption{Instruct directives for Prompt and Reward Models.}
    \label{tab:instruction_sets}

\end{minipage}

\end{table*}

\end{document}